%% file: main.tex
\documentclass[sigconf]{acmart}

\usepackage{graphicx}
\usepackage{epsfig}
\usepackage{booktabs} 
\usepackage{multirow}
\usepackage{amssymb}
\usepackage{enumitem}
\usepackage{subcaption}
\usepackage{paralist}

\def \x {\mathbf{x}}

\makeatletter
\newcommand{\thickhline}{%
    \noalign {\ifnum 0=`}\fi \hrule height 1pt
    \futurelet \reserved@a \@xhline
}
\newcommand*{\centerfloat}{%
  \parindent \z@
  \leftskip \z@ \@plus 1fil \@minus \textwidth
  \rightskip\leftskip
  \parfillskip \z@skip}
\makeatother

\setcopyright{acmlicensed}

\acmDOI{}

\acmISBN{}

\acmConference[LAK'19]{International Learning Analytics and Knowledge Conference}{March 2019}{Tempe, Arizona}
\acmYear{2019}
\copyrightyear{2019}

\acmArticle{}
\acmPrice{}

\editor{}

\begin{document}
\title[Effective Feature Learning with Unsupervised Learning in MOOCs]{Effective Feature Learning with Unsupervised Learning for Improving the Predictive Models in Massive Open Online Courses}

\author{Mucong Ding}
\affiliation{%
  \institution{Department of Computer Science and Engineering\\Hong Kong University of Science and Technology\\Hong Kong SAR, China}
}
\email{mcding@connect.ust.hk}

\author{Kai Yang}
\affiliation{%
  \institution{Department of Computer Science and Engineering\\Hong Kong University of Science and Technology\\Hong Kong SAR, China}
}
\email{yangkai@cse.ust.hk}

\author{Dit-Yan Yeung}
\affiliation{%
  \institution{Department of Computer Science and Engineering\\Hong Kong University of Science and Technology\\Hong Kong SAR, China}
}
\email{dyyeung@cse.ust.hk}

\author{Ting-Chuen Pong}
\affiliation{%
  \institution{Department of Computer Science and Engineering\\Hong Kong University of Science and Technology\\Hong Kong SAR, China}
}
\email{tcpong@cse.ust.hk}

\renewcommand{\shortauthors}{M. Ding et al.}

\begin{abstract}
The effectiveness of learning in massive open online courses (MOOCs) can be significantly enhanced by introducing personalized intervention schemes which rely on building predictive models of student learning behaviors such as some engagement or performance indicators. A major challenge that has to be addressed when building such models is to design handcrafted features that are effective for the prediction task at hand. In this paper, we make the first attempt to solve the feature learning problem by taking the unsupervised learning approach to learn a compact representation of the raw features with a large degree of redundancy. Specifically, in order to capture the underlying learning patterns in the content domain and the temporal nature of the clickstream data, we train a modified auto-encoder (AE) combined with the long short-term memory (LSTM) network to obtain a fixed-length embedding for each input sequence. When compared with the original features, the new features that correspond to the embedding obtained by the modified LSTM-AE are not only more parsimonious but also more discriminative for our prediction task. Using simple supervised learning models, the learned features can improve the prediction accuracy by up to 17\% compared with the supervised neural networks and reduce overfitting to the dominant low-performing group of students, specifically in the task of predicting students' performance. Our approach is generic in the sense that it is not restricted to a specific supervised learning model nor a specific prediction task for MOOC learning analytics.
\end{abstract}

%
%



\maketitle

\input{body}

\bibliographystyle{ACM-Reference-Format}
\bibliography{references}
\end{document}

%% file: body.tex
\section{Introduction}
With the advancement in digital technology, massive open online courses (MOOCs) have become popular over the past decade and provide alternative ways of learning. With the current one-size-fits-all approach of MOOCs, the students need to be self-motivated and self-disciplined throughout the whole learning period without much individual guidance from the instructor, and as a consequence, many are prone to drop out of courses \cite{daradoumis2013review}. In terms of the high attrition rate in MOOCs, there has been a great deal of interest in providing personalized learning guidance with the help of predictive models, to improve students' motivation and the effectiveness of learning which should result in an increased retention rate \cite{whitehill2015beyond}. The predictive models proposed in the literature \cite{chaplot2015predicting, kloft2014predicting, fei2015temporal}, help depict the learning progress, achieve a better understanding of the learning abilities, identify the at-risk students, and provide proactive interventions. However, when building such models with supervised learning, it is still challenging to design handcrafted features that characterize students' learning behaviors such as some engagement or performance indicators, and are effective for the prediction tasks. 

In MOOCs, students' learning behavior is complicated in both the time domain and the content domain (the sequence of how course materials are ordered). An instructor usually designs the course schedule according to the knowledge graph which depicts the inherent dependence and hierarchical structure among course materials. The content-based representation is important, as the course materials are what a student interacts with, and the information of the course materials can only be encoded within the content domain. However, the learning pattern hidden in the content domain is usually missed in the current research, as all the models of the prevalent dropout prediction \cite{chaplot2015predicting, kloft2014predicting, fei2015temporal, whitehill2017mooc, halawa2014dropout, he2015identifying, ye2014early} are devised to capture the learning pattern in the time domain. If only the temporal features are utilized, dropout prediction can only provide preliminary interventions such as sending email alerts to at-risk students, let alone providing effective pedagogic advice and support with regard to the course materials for each student to improve the effectiveness of learning. Actually, the MOOC platform allows students' learning activities including video watching, page navigation, quiz participation to be recorded as the clickstream logs. Each clickstream is a collection of records, while each record consists of student ID, interaction time and accessed course materials so that we can extract both time and content domain features from the clickstream. Because of the strong inter-dependencies between the course materials with close locations in the knowledge graph, the content-domain features exhibit contextual locality, i.e., features with close locations in the sequence are more correlated. This enables us to utilize the underlying local patterns in the content domain for a variety of prediction problems.

The knowledge provided in the course content is the most important part of MOOCs. When attaining the content mastery of a specific concept, a student needs to go to the relevant pages, watch videos and answer questions with a sequence of actions recorded in the clickstream. Therefore, in terms of the many interactions related to knowledge mastery, there is a large degree of information redundancy in the clickstream data, so that it is challenging to design a set of effective handcrafted features to discriminate between the learning patterns of the high and low performing students using the raw information in the clickstream data. Feature learning with unsupervised models is a state-of-the-art approach to this problem \cite{boschunsupervised}. As it is capable of learning an effective and compact representation of the raw features, which not only simplifies the architectures of the predictive models but also helps improve the prediction performance.

In this paper, we prepare the features in the time-content domain and make the first attempt to solve the feature learning problem by taking the unsupervised learning approach to learning a compact and effective representation from the highly redundant information in the raw records. Specifically, we design a modified auto-encoder (AE) combined with the long short-term memory (LSTM) networks to learn the new features that correspond to the representation of the embedding layer. The key points of this paper are summarized as follows: 
\begin{asparaenum}
\item We prepare raw features and analyze the learning behavior in the time-content domain.
\item We show that neural networks outperform logistic regressions by a large margin in predicting the next chapter grade.
\item We propose a modified auto-encoder (AE) combined with the long short-term memory (LSTM) network (\textbf{modified LSTM-AE}) and two variational auto-encoders (VAEs) to learn compact and effective representations from the raw features.
\item Compared with the VAEs, the representation in \textbf{modified LSTM-AE} is better at discriminating the low and high performing students.
\item With the \textbf{modified LSTM-AE}, the learned features help reduce the overfitting to the dominant low-performing group of students and improve the performance in the specific task of predicting the next chapter grade by up to 17\% compared with the completely supervised neural network baselines.
\end{asparaenum}

\section{Background}
In this section, we introduce how the course materials are managed in edX, and how to formulate the learning performance prediction problem as a supervised learning task. Then, we introduce logistic regression and design some specific neural networks as the baselines to predict students' learning performance. We analyze the high-degree of redundancy problem of handcrafted features and point out that representation learning is a potent solution. Finally, we discuss restrictions of real-time predictions, i.e., predicting alongside with the progression of a course.

\subsection{\label{sec_activity_and_assessment}Learning Activity and Assessment}
The materials of a MOOC in edX are managed in a hierarchical way: \{~{\em videos}, ~{\em problems}, ~{\em others} \} $\in$ ~{\em verticals} $\in$ ~{\em sequentials} $\in$ ~{\em chapters} $\in$ ~{\em course}, where a sequential corresponds to a subsection in a chapter. In each sequential, a set of problems, videos, and other course materials are listed in the verticals. A set of assignments in the problem verticals constitute the assessments of a sequential, and the sequential grade is accordingly calculated by aggregating the weighted scores of the problem verticals according to the grading policy. The chapter grades and the course grade can be obtained similarly. The numerical grades obtained indicate the students' mastery levels of the knowledge associated with the course content.

\begin{table*}[t]
    \centering
    \caption{\label{table:feature_list}List of Features}
\begin{tabular}{cll}
\thickhline
\multicolumn{2}{c}{\textbf{Feature}}                   & \multicolumn{1}{c}{\textbf{Explanation (in the content domain)}}  \\ \hline
\multirow{2}{*}{\textbf{Navigation}} & navigate-forward  & Number of forward navigation events for a specific assignment     \\
                                     & navigate-backward & Number of backward navigation events for a specific assignment    \\ \hline
\multirow{8}{*}{\textbf{Video}}      & load-video        & Number of loading video events for a specific assignment          \\
                                     & play-video        & Number of playing video events for a specific assignment          \\
                                     & pause-video       & Number of pausing video events for a specific assignment          \\
                                     & stop-video        & Number of stopping video events for a specific assignment         \\
                                     & seek-backward     & Number of seeking video backward events for a specific assignment \\
                                     & seek-forward      & Number of seeking video forward events for a specific assignment  \\
                                     & show-subtitle     & Number of showing subtitle events for a specific assignment       \\
                                     & hide-subtitle     & Number of hiding subtitle events for a specific assignment        \\ \hline
\end{tabular}
\end{table*}

\begin{figure}[t]
    \centerfloat
    \includegraphics[width=1.1\linewidth]{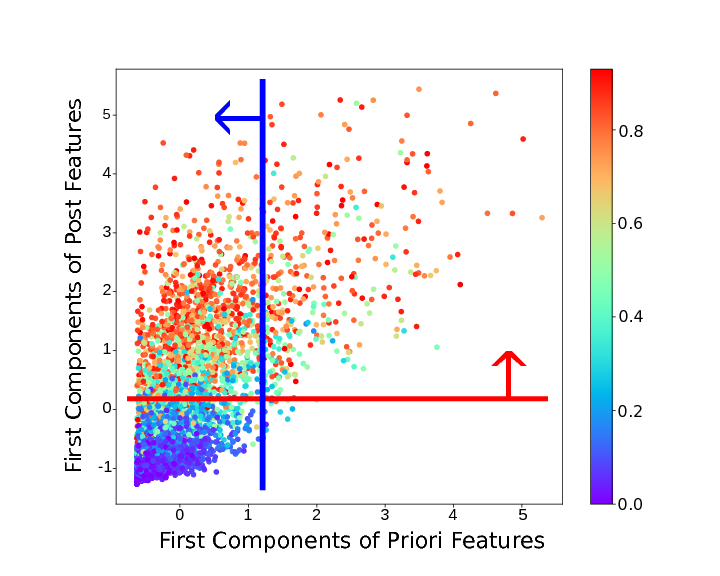}
    \caption{\label{fig:prior_post_assignment_scatter} Principal component analysis (PCA) on the prior and post assignment features, which capture student activities before and after the time when the student completed the corresponding assignment respectively. The color shows each student's average chapter grade.}
\end{figure}

In the content domain, video and navigation activities are the two most important learning behaviors and activities related to knowledge mastery. In this paper, we extract several raw features which aim to characterize specific behaviors related to video and navigation from the clickstream data, and summarize them in Table~\ref{table:feature_list}. Each feature is two-fold in the time domain to capture activities before and after the split time when the student completed the corresponding assignment. For example, the feature \texttt{load-video} is split into \texttt{load-video-prior} and \texttt{load-video-post}, where the former is the number of loading video events for a assignment before its submission, and latter is corresponding count after the submission. This two-fold separation in the time domain is useful for capturing different behaviors of students with different performances. As shown in Figure~\ref{fig:prior_post_assignment_scatter}, the high and low performing students show different behaviors before and after the completion of assignments. Students with low performance (blue dots) have not completed enough prior-quiz studies (as they are located to the left of the blue line, their prior features are relatively small), while students with high grades (red dots) have at least completed some post-quiz studies (as they are located above the red line, their post features are large). As described above, we attempt to prepare the features in the content-time domain, with the two-fold representation in the time domain.

The learning progress of a student in edX can be organized as sequences. Suppose there are $N$ sets of course materials, all of which have the corresponding assessments, whose order in the course is pre-defined. For a specific student, we define the sequence of learning behaviors $(\x_1, \cdots , \x_N)$ as the raw features, where $\x_i\in\mathbb{R}^F$ is a list of features characterizing his activities in the $i$-th set of materials, and define the sequence of grades $(y_1,...,y_N)$ as the labels. In this paper, we prepare the set of course materials at the chapter level, and $N$ is the total number of chapters. The sequence of assignment grades is not independent owing to the prerequisite dependency in the knowledge graph. A student who gets a good grade in the current assessment is likely to do well in the future assessments.

\subsection{Formulation of the Performance Prediction Problem}
\begin{figure}[t]
    \centerfloat
    \includegraphics[width=0.8\linewidth]{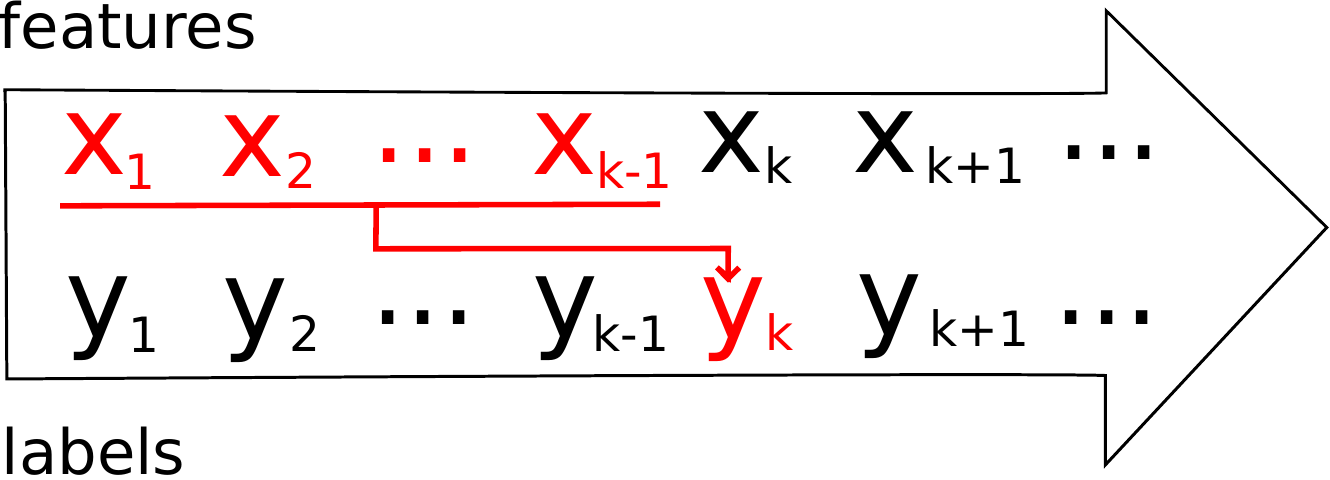}
    \caption{\label{fig:performance_prediction_formulation} Formulation of the performance prediction problem. For the predicted grade in chapter $k$, only features up to chapter $k-1$ are available.}
\end{figure}

The performance prediction is essentially a sequence labeling problem, which uses the raw features of a student before a specific chapter $k$, i.e. $[\x_1, \x_2, \dots, \x_{k-1}]$, to predict the student's grade in the next chapter, i.e. $y_k$, where $1 < k \leq N$, as shown in Figure~\ref{fig:performance_prediction_formulation}. We train a list of $(N-1)$ independent models, $[f_2, \dots, f_{N}]$, one at a time, so that $f_k$ specifically predicts the grades of chapter $k$. The loss function of model $f_k$ for a specific sample is defined as $\ell = (y_k-\hat{y}_k)^2$ where $\hat y_k$ is the predicted chapter grade and $y_k$ is the ground truth label.

\subsection{\label{sec_prediction_baseline}The Baselines of Performance Prediction}
\subsubsection{Logistic Regression}
In statistics, the logistic model is a statistical model with input (independent variable) a continuous variable and output (dependent variable) a binary variable, where a unit change in the input multiplies the odds of the two possible outputs by a constant factor. Logistic regression (\textbf{LR}) is widely used for binary classification, and we consider it as a baseline model to evaluate the performance of neural networks.

\subsubsection{Multi-layer Perceptron}
The multi-layer perceptron (MLP) is a class of artificial neural networks based on a collection of units called artificial neurons, an analogous to biological neurons in an animal brain. With the neuron connection, MLP focuses on approximating the complex relationship between the input and output. An MLP consists of multiple fully connected layers of neurons and is able to learn the dependency between a collection of distinguishing features and the target labels. In this paper, we design the fully connected networks with three hidden layers (\textbf{FC3}) as a baseline.

\subsubsection{Convolution Neural Networks}
The convolutional neural network (CNN) applies a group of neurons (a kernel) across a specific dimension of the data. The neurons thus are capable of learning features that are defined by local patterns possibly occurring anywhere in the input. When the data is heterogeneous, it is meaningless to apply the convolutional filter along a list of distinguishing features. For our content-domain features, because the order of course materials preserves the implicit dependence on the knowledge graph, we can apply one-dimensional convolution to the content domain to find the implicit learning patterns. For most of the MOOCs, since the number of chapters is limited to $N\leq12$, we always set the kernel size to $3$. We design a neural network with two convolutional layers followed by a fully connected layer (\textbf{CNN2-FC1}) as a baseline.

\subsubsection{Long Short-Term Memory Networks}
Recurrent neural networks (RNNs) are a class of network structures in which the neurons at each layer are not only connected to neurons of adjacent layers but also receive the input from the same layer at the previous step in a sequence. With these connections across steps in sequential data, RNN is capable of learning patterns that change dynamically over that dimension. The RNNs with long short-term memory (LSTM) cells are used to overcome the gradient vanishing problem in the vanilla RNN and is now a widely used standard in sequence learning problems. In our case, LSTM networks are potent to capture the hidden learning patterns within the content domain. Here, we design a neural network with one LSTM layer (\textbf{LSTM1}), and another model consists of a one-dimensional CNN of kernel size $1$ followed by an LSTM layer (\textbf{CNN1-LSTM1}) as baselines.

\subsection{Feature Redundancy and Representation Learning}
\begin{figure}[t]
    \centerfloat
    \includegraphics[width=1.1\linewidth]{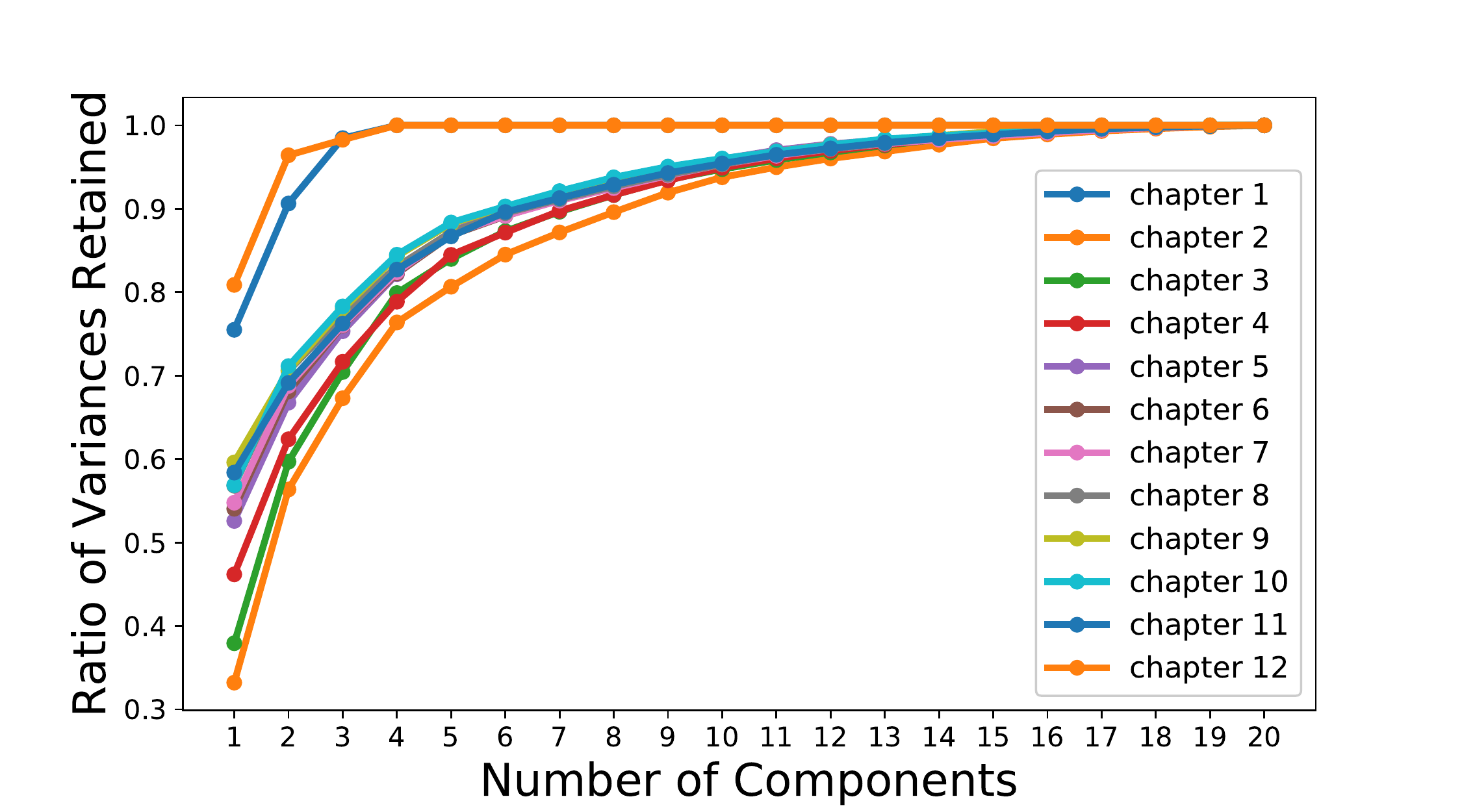}
    \caption{\label{fig:retained_variance_ratio} The dependency of the ratio of variances retained on the number of output components with the PCA for each chapter.}
\end{figure}

Feature redundancy is a common problem of hand-crafted features in the context of MOOC data analytics. Even for the simple purpose of learning specific content, students need to go through a collection of relevant pages (e.g., watching videos and answering quizzes), which triggers a sequence of interdependent actions recorded in the clickstream. In this regard, the underlying correlations might be significant among different types of interaction events, and the manually extracted features could be highly redundant. This gives rise to the increasing difficulty of training a robust predictive model. For the specific set of features we extracted in this paper, the redundancy problem can be shown by performing principal component analysis (PCA) directly on the 20 features of a specific chapter $k$.  From Figure~\ref{fig:retained_variance_ratio} we can see that 5 components can retain over 80\% of the variances for all chapters.

Unsupervised learning using auto-encoders (AEs) for finding a distinctive representation of the raw features is a potential solution to the feature redundancy problem. Since our features and labels are essentially sequences, the auto-encoding of features is a sequence-to-sequence mapping problem. And thus we should search for an unsupervised architecture which is capable of extracting the underlying representations of general sequences. Outside of the educational context, a general sequence-to-sequence learning framework is used in natural language processing and video representation learning, where a long short-term memory (LSTM) network is used to encode a sequence into a fixed-length representation, and then another LSTM network is used to decode a sequence out of that representation \cite{sutskever2014sequence}. Following this well-established scheme, we explore AEs combined with LSTM networks to obtain a compact embedding of MOOC data in the educational context, which also aims to be efficient for prediction problems.

\section{Representation Learning}
In this section, we propose our modified LSTM auto-encoder for representation learning. We also list two other designs of variational auto-encoders (VAEs) with the symmetric and asymmetric structures for comparison purposes.

\subsection{The Modified LSTM Auto-encoder}
Inspired by the unsupervised LSTM auto-encoder which has achieved a big success in video representation learning \cite{pmlr-v37-srivastava15}, we propose a modified LSTM auto-encoder (denoted by \textbf{Modified LSTM-AE}) for learning efficient and compact representations of feature sequences in the context of MOOC data analytics.

\begin{figure}[t]
    \centerfloat
    \includegraphics[height=\linewidth, angle=90]{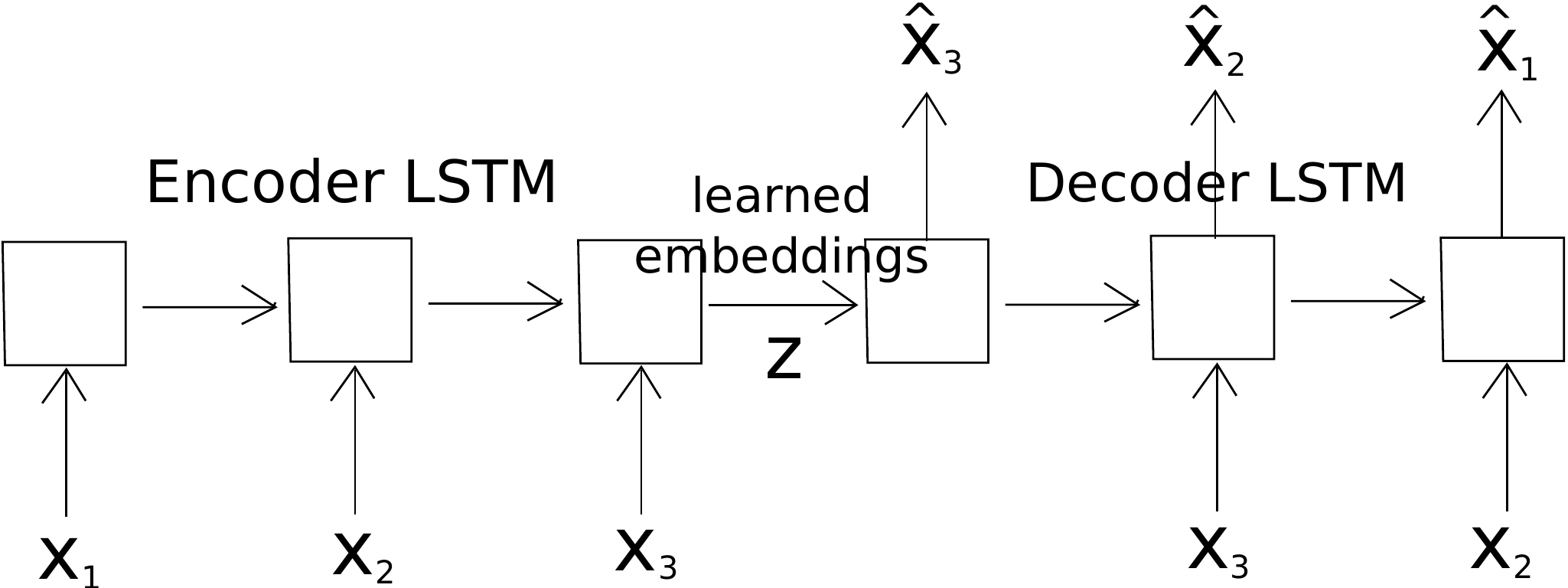}
    \caption{\label{fig:lstm_ae_architecture} Architecture of the LSTM AE when $k=4$.}
\end{figure}

We first describe the LSTM auto-encoder model, which combines the well-established sequence-to-sequence learning framework \cite{sutskever2014sequence} and the idea of using auto-encoders for representation learning. It consists of two LSTM networks, the encoder, and the decoder LSTM, as shown in Figure~\ref{fig:lstm_ae_architecture}. The features in the input sequence $(\x_1, \x_2, \dots, \x_{k-1})$ are fed into the encoder LSTM one at a time. Right after the last input $\x_{k-1}$ has been read, the last output from the encoder, $\boldsymbol{z}$, is the learned embedding and is forwarded to the decoder LSTM as the first input. The decoder then reads in the original sequence (except from $\x_1$) in reverse order, $(\x_{k-1}, \dots, \x_2)$. Its output $(\hat{\x}_{k-1}, \hat{\x}_{k-2}, \cdots, \hat{\x}_{1})$ aims to reconstruct the input sequence. Because of the contextual locality of the content domain (features with close locations in the sequence are more correlated), it is much easier for the decoder to reconstruct the input sequence in reverse order. And hence the entire LSTM auto-encoder adopts the last-in-first-out (LIFO) scheme. Note that no matter how long the input sequence is, the learned representation has a fixed length, which prohibits the auto-encoder to learn a trivial identity mapping.

We could think of our LSTM auto-encoder as first recursively performing an LSTM operation (defined by the weights of the encoder LSTM) to encode, and then recursively performing another LSTM operation to decode. In this sense, features close to the turning point (chapter $k$) has the potential to dominate the learned embedding, as some features in the early chapters decay exponentially during the recursive encoding. This characteristic helps the LSTM auto-encoder learn an efficient representation for real-time predictions (predicting labels in the next chapter) tasks. We expect features close to chapter $k$ are more useful in prediction because of the contextual locality, while the local patterns in early sequences could be meaningless. Note that the decoder is conditioned on the ground truth features (i.e., we input the ground truth features to the decoder again), which enables this LSTM auto-encoder to capture multiple modes in a sequence.

The real-time framework does not forbid us to input the feature sequences after chapter $k$ into the decoder part (see Sec.\ref{sec_real_time} for our detailed argument). Thus a notable short-coming of the LSTM auto-encoder is that it does not utilize this subsequence of ground truth features, $(\x_{k}, \x_{k+1}, \x_{k+2}, \cdots, \x_N)$, to enhance the predictive power of the embedding. As we merely require the model to reconstruct the input sequence, the learned representation only contains information collected before chapter $k$, and this is the main restriction of the effectiveness of our learned representation. 

\begin{figure}[t]
    \centerfloat
    \includegraphics[height=\linewidth, angle=90]{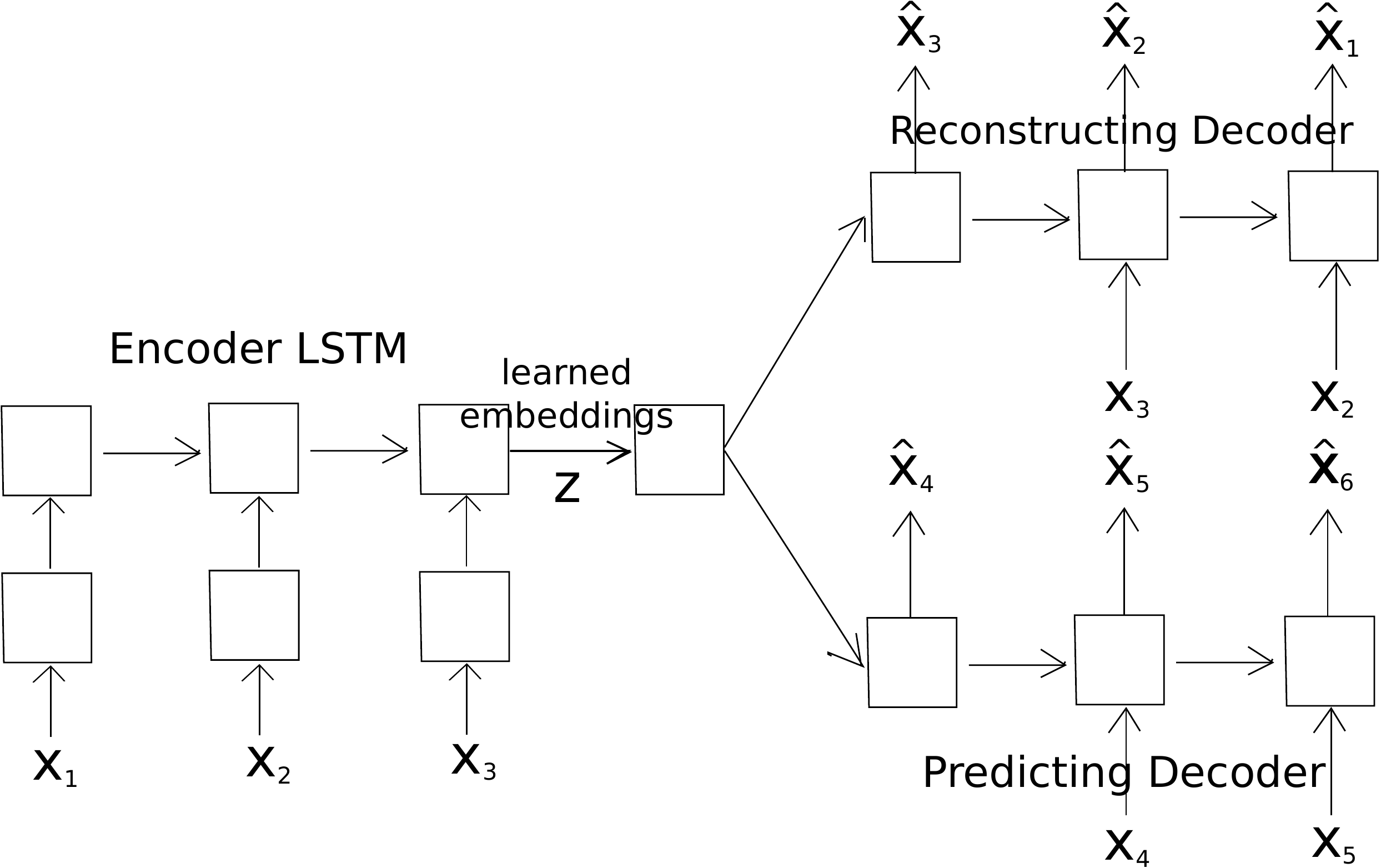}
    \caption{\label{fig:modified_lstm_ae_architecture}Architecture of the \textbf{Modified LSTM-AE} model when $k=4$ and $N=6$.}
\end{figure}

We could improve the design by adding another predicting decoder parallel to the original reconstructing decoder, which aims to predict the sequences of features in the later chapters at the same time (see Figure~\ref{fig:modified_lstm_ae_architecture}). If this decoder could predict the features in the next few chapters correctly, the encoder must have captured some latent patterns of the input sequences, and such information is contained in the learned embedding. This prohibits the encoder to ignore all of the useful information in the early part of the sequence. Moreover, during the training process, some information from the subsequence following chapter $k$ could be captured by the weights of encoder LSTM though stochastic gradient descent based optimization. Hence, the embedding can find useful patterns from the entire sequence now, even though we only input the subsequence $(\x_1, \x_2, \dots, \x_{k-1})$ to the encoder. This greatly eases the difficulty of predicting with the embedding, since our encoder now has some prediction capability as well. Similarly, we also provide the ground truth features to the reconstructing decoder to help it extract multiple modes from the sequence.

One small defect of our current design is that our embedding, $\boldsymbol{z}$, must have exactly the same size as the feature units, $\boldsymbol{z}\in\mathbb{R}^Z$ and $Z=F$, otherwise the decoders cannot take both of them as input. This is a restriction that we want to get rid of. Actually, we can further shrink the size of the embedding space by applying a one-dimensional convolutional layer of kernel size 1 before the LSTM layer in the encoder. This layer further reduces the dimensionality of the inputs before extracting the underlying patterns. At the same time, we add one fully connected layer before the two decoders to map the small embedding space $\mathbb{R}^Z$ to $\mathbb{R}^F$, where $Z<F$. The complete design is shown in Figure~\ref{fig:modified_lstm_ae_architecture}. Note that this architecture is also capable of enlarging the embedding space to $\mathbb{R}^Z$ where $Z>F$ under the situation where $F$ is too small for learning a good representation.

The loss of this \textbf{Modified LSTM-AE} model is generally defined as the weighted mean squared error (MSE) between the complete input sequence $[\x_1, \x_2, \dots, \x_N]$ and the output $[\hat{\x}_1, \hat{\x}_2, \dots, \hat{\x}_N]$. In order to further encourage our model to learn a representation which utilizes the contextual locality around the turning point, we assign a set of Gaussian weights $\mathrm{exp}[(k-n)^2/2\sigma^2]$ to each of the MSEs between the pair $(\x_n, \hat{\x}_n)$, $(1\leq n\leq N)$, where $\sigma$ is usually set to $3$. For a single sample, the loss is defined as $\ell=\sum_{n=1}^N\mathrm{exp}[(k-n)^2/2\sigma^2](\x_n-\hat{\x}_n)^2$. You can see that we assign larger weights to MSEs around chapter $k$. This forces our model to treat learning the local representation as the first priority.

\subsection{The Baselines of Representation Learning}
\begin{figure}[t]
    \centerfloat
    \includegraphics[height=0.45\linewidth, angle=90]{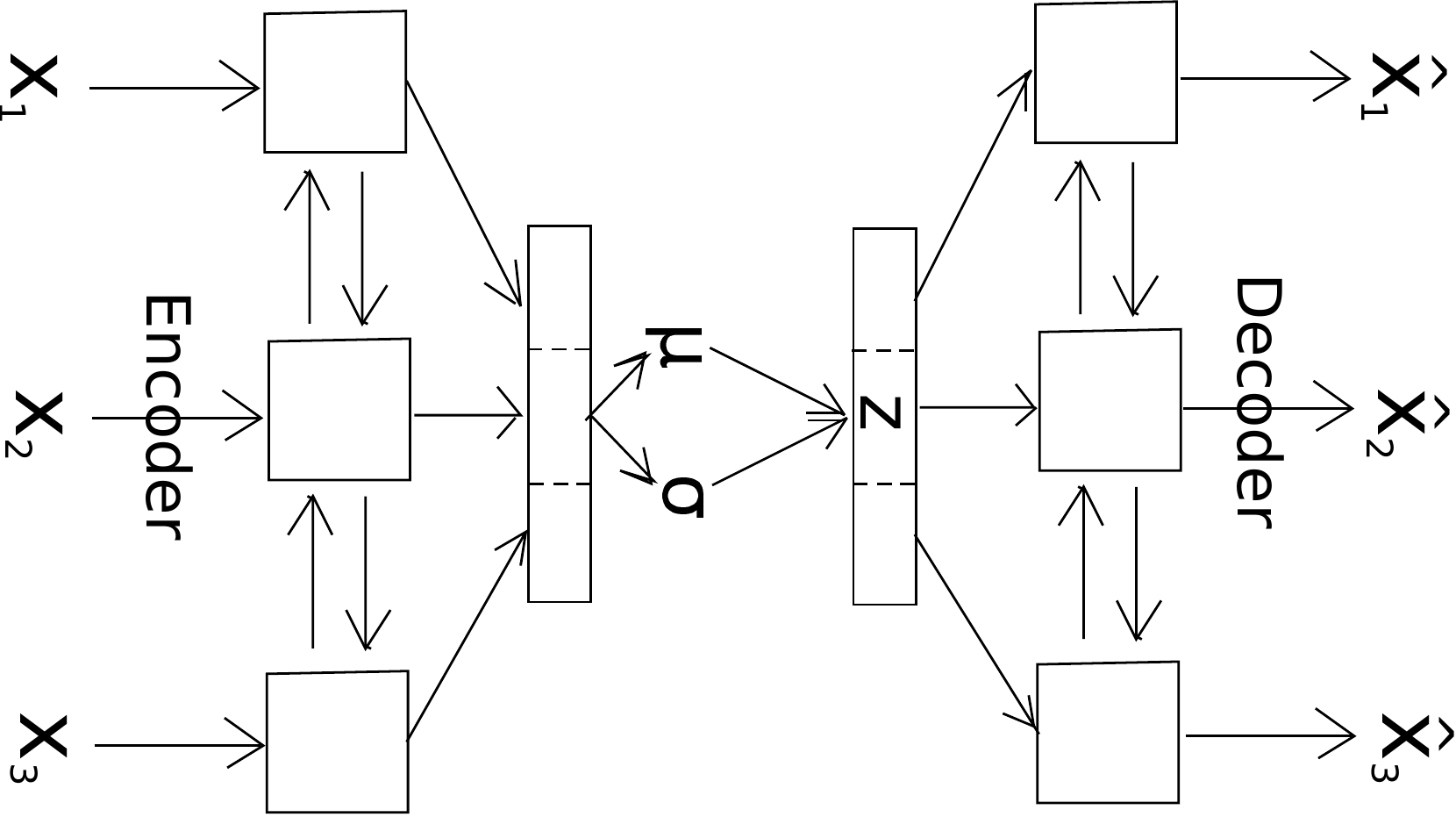}
    \caption{\label{fig:symmetric_vae_architecture} Architecture of the \textbf{Symmetric-VAE} model when $k=4$.}
\end{figure}

\begin{figure}[t]
    \centerfloat
    \includegraphics[height=0.5\linewidth, angle=90]{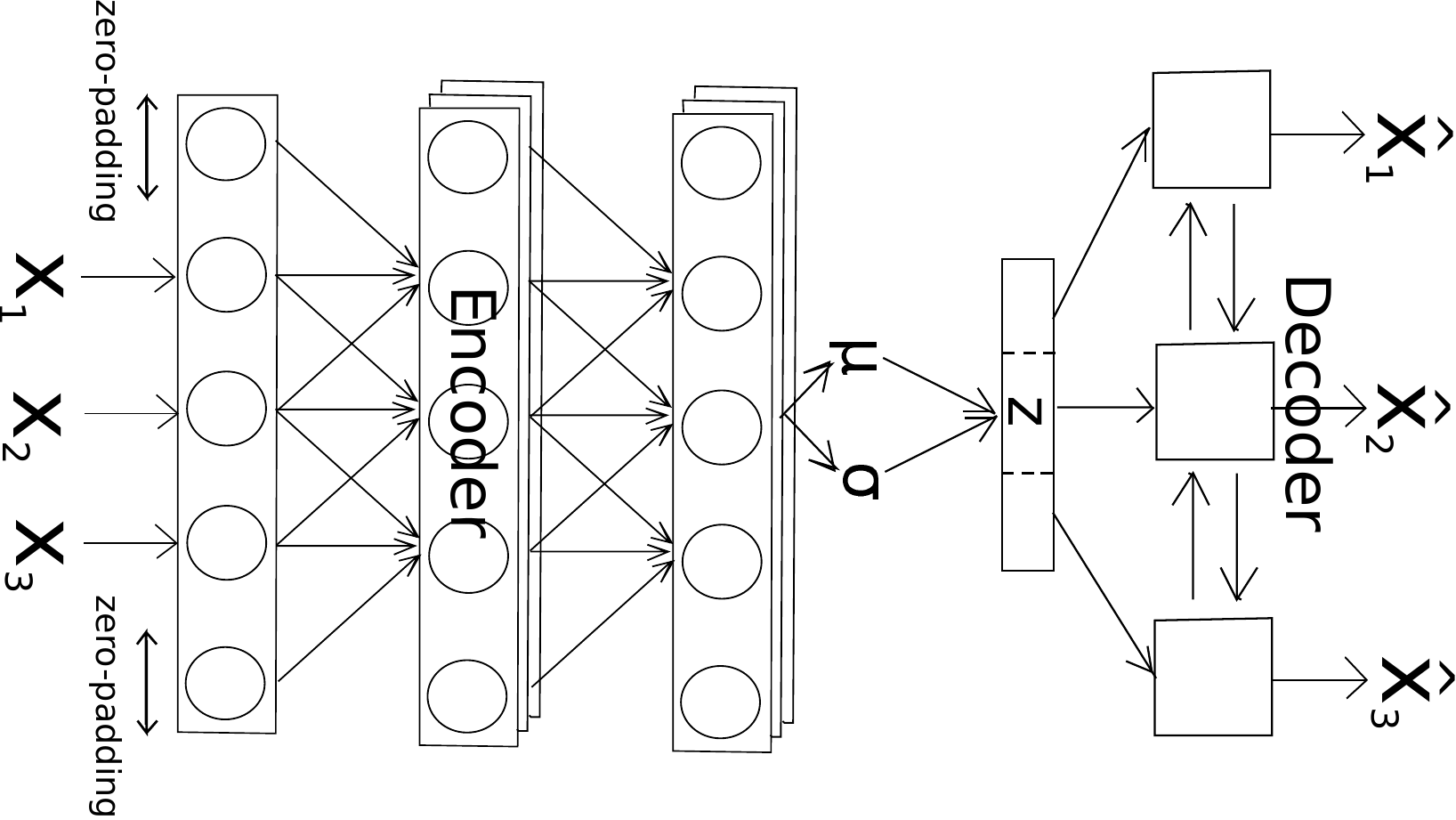}
    \caption{\label{fig:asymmetric_vae_architecture} Architecture of the \textbf{Asymmetric-VAE} model when $k=4$.}
\end{figure}

Variational auto-encoder (VAE) is a popular approach to learning robust embeddings for various types of data. Here, we build two VAEs as baselines to compare with the proposed \textbf{Modified LSTM-AE}. The first VAE we build (denoted by \textbf{Symmetric-VAE}) processes a symmetric structure \cite{boschunsupervised}. As shown in Figure~\ref{fig:symmetric_vae_architecture}, both the encoder and the decoder of it is one layer of the bi-directional LSTM network. Where in a bi-directional LSTM, we connect two hidden layers of opposite directions to the same output. Thus future input information is reachable from the current state.  the another VAE, the encoder consists of three one-dimensional convolutional layers (Figure~\ref{fig:asymmetric_vae_architecture}). We call it \textbf{Asymmetric-VAE}, since it has an asymmetric structure and is capable of extracting more low-level feature embeddings \cite{klingler2017efficient}.

The biggest difference between the VAEs and our \textbf{Modified LSTM-AE} model is that they do not have a fixed-length embedding. In the VAEs, the content domain is preserved in the embedding layer with the same sequence length. This approach has the advantage of capturing useful patterns anywhere in the feature sequence and makes the reconstruction of the original input easier. However, the embeddings learned by those VAEs may contain many useless local patterns in the content domain which are relatively far from the chapter we are predicting. In this sense, we argue that the two VAEs are not as powerful as the \textbf{Modified LSTM-AE} model in finding efficient representations for real-time predictions. This is what we see in the experiments.

\subsection{Prediction with Learned Representations}
Equipped with the unsupervised model which learns a good representation of the feature sequences, we could build a relatively simple predictor which takes the learned representations as the new feature inputs. For the \textbf{Modified LSTM-AE} model, we use a fully connected network with one hidden layer as the predictor. For the two VAEs, since the embedding is still a sequence, we use one layer of LSTM network as the predictor.

When training these predictors, it is possible to fine-tune the weights of the pre-trained encoders to enhance the representation for a specific prediction task further. After fine-tuning, the portion of useful features for this specific task in the embedding space could be further enlarged. During such a training process, a much smaller learning rate should be used for the encoder part, as we expect the pre-trained encoder can already find a very efficient representation for a variety of prediction problems.

\subsection{\label{sec_real_time}Real-time Interventions Based on the Predictive Model}
Prediction along with the progression of a course (i.e., predicting students' behaviors in the next chapter $k$ when they are learning chapter $k-1$) is crucial for real-time interventions. This one-step ahead foresight can provide students with the effective pedagogic support on the current course material at the right time. However, real-time prediction also imposes a strong restriction on the training setups, in the sense that only features and labels up till chapter $k-1$ of the current course are available to us when we are making predictions for chapter $k$. As a consequence, the typical training procedure which requires labels defined in chapter $k$ does not work on a course in progress. Instead, we have to learn the model retrospectively on data generated from a finished course \cite{boyer2015transfer}. Under this transfer learning scenario, although all the data is actually available, when we train a predictor on a completed course, we are not allowed to input features in or after chapter $k$ to the model, because we cannot do the same thing when making predictions using this trained predictor on an ongoing course. However, when we train an unsupervised model on a finished course, we can input the ground truth features in or later than chapter $k$ to the decoder part (like what we do in \textbf{Modified LSTM-AE}), as long as its encoder part only reads in the set of features before chapter $k$. This is because only the encoder which generates the learned embedding is used when making predictions, and as long as we do not make use of the features in and after week $k$ when predicting on an ongoing course, we do not break the causality.

\begin{figure*}[t]
    \centerfloat
    \begin{subfigure}[c]{0.5\linewidth}
        \includegraphics[width=\linewidth]{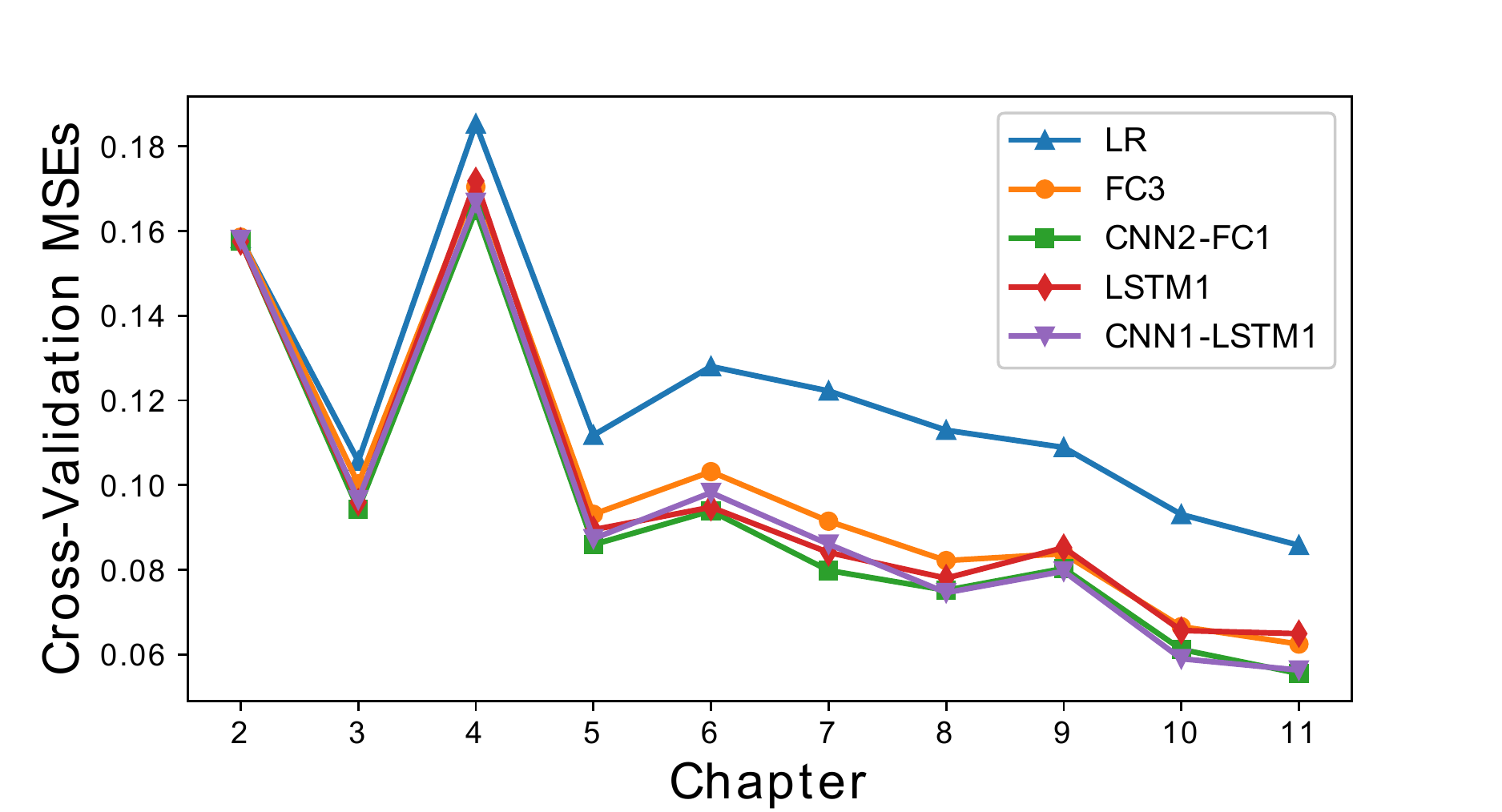}
    \end{subfigure}
    \begin{subfigure}[c]{0.5\linewidth}
        \includegraphics[width=\linewidth]{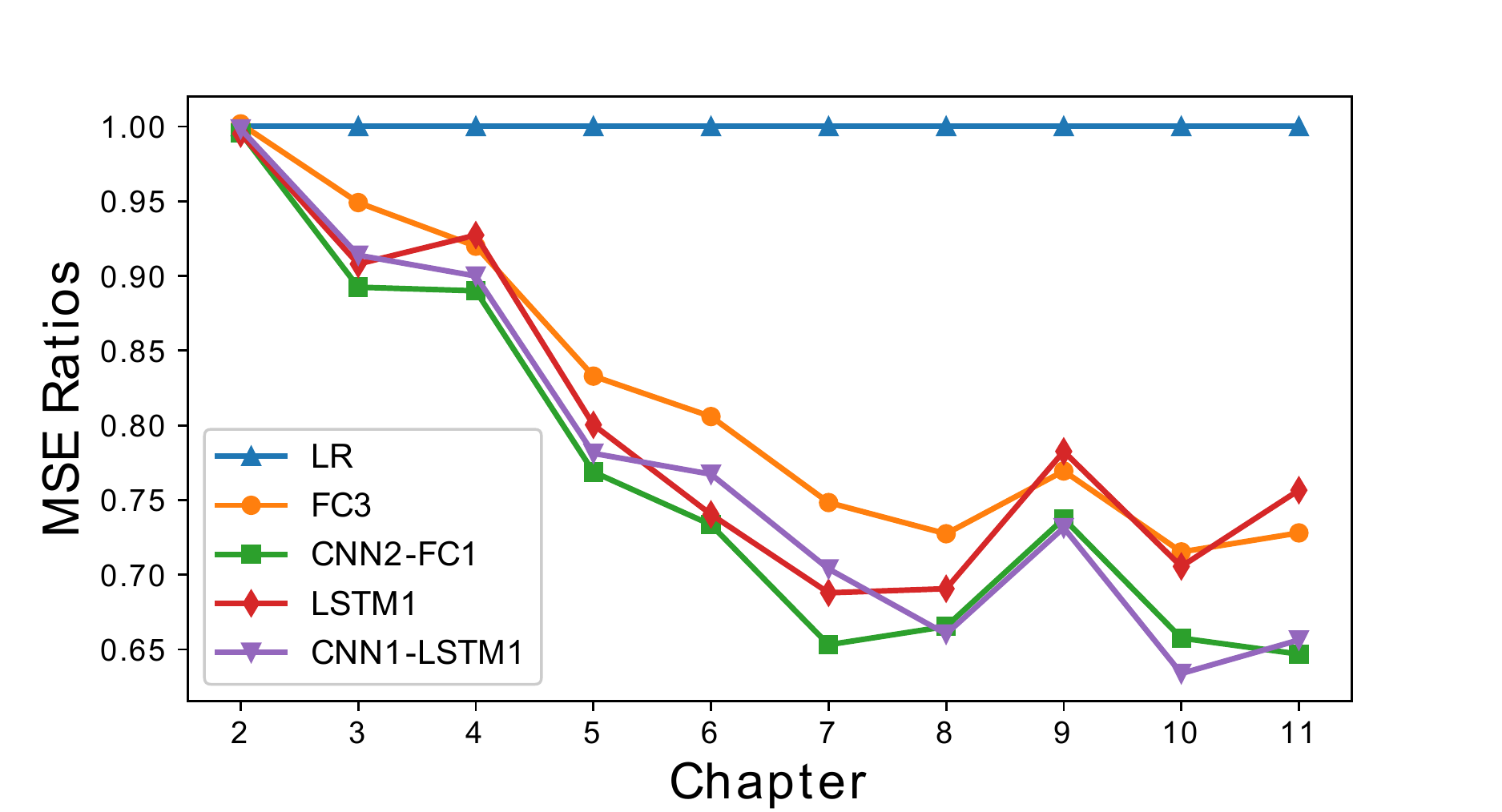}
    \end{subfigure}
    \caption{\label{fig:predictor_performance}The left figure shows the comparison of prediction performance among supervised baselines. The right figure shows the relative prediction performance compared to logistic regression (\textbf{LR}).}
\end{figure*}

\begin{figure*}[t]
    \centerfloat
    \begin{subfigure}[c]{0.54\linewidth}
        \includegraphics[width=\linewidth]{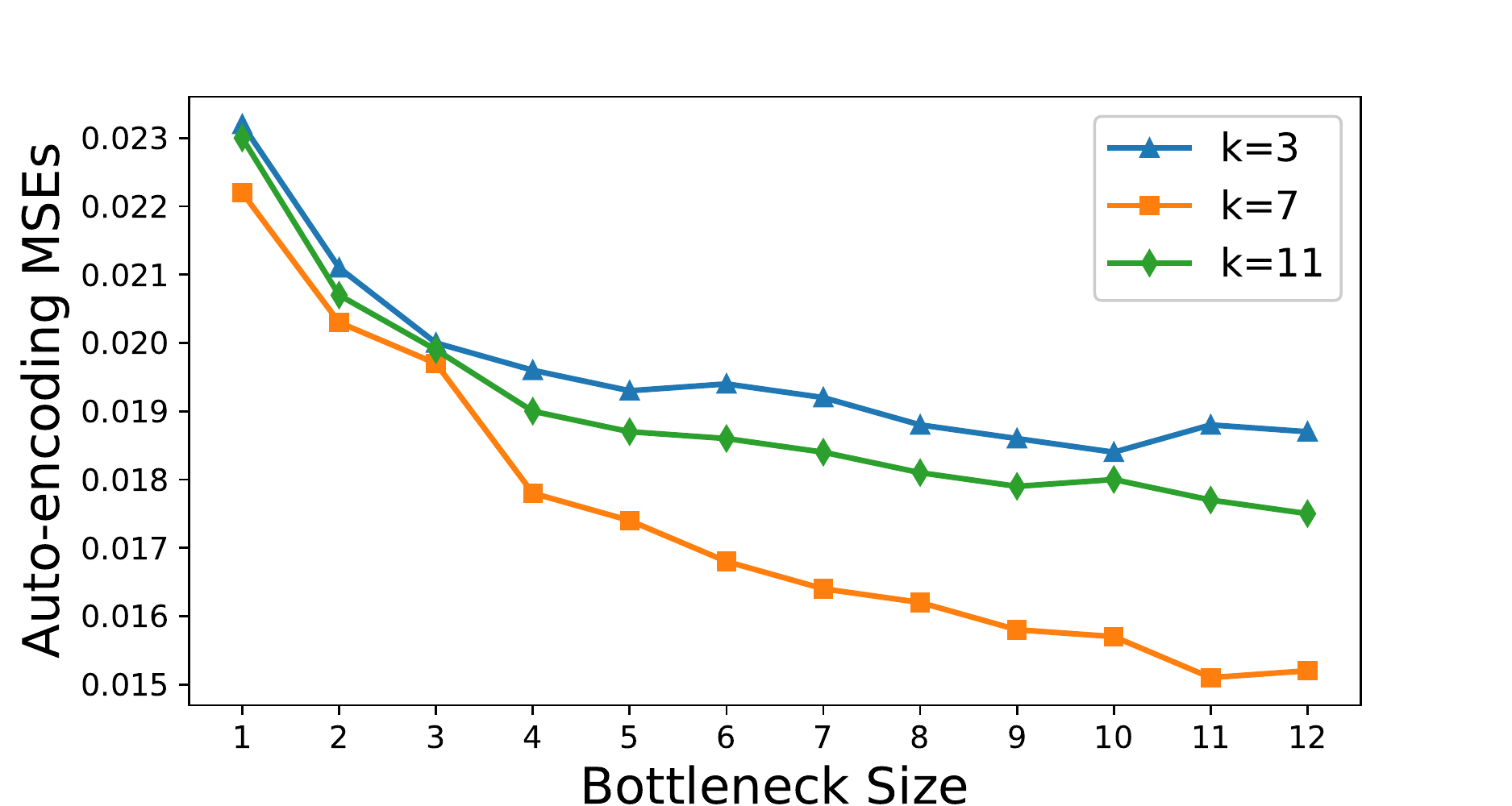}
    \end{subfigure}
    \begin{subfigure}[c]{0.43\linewidth}
        \includegraphics[width=\linewidth]{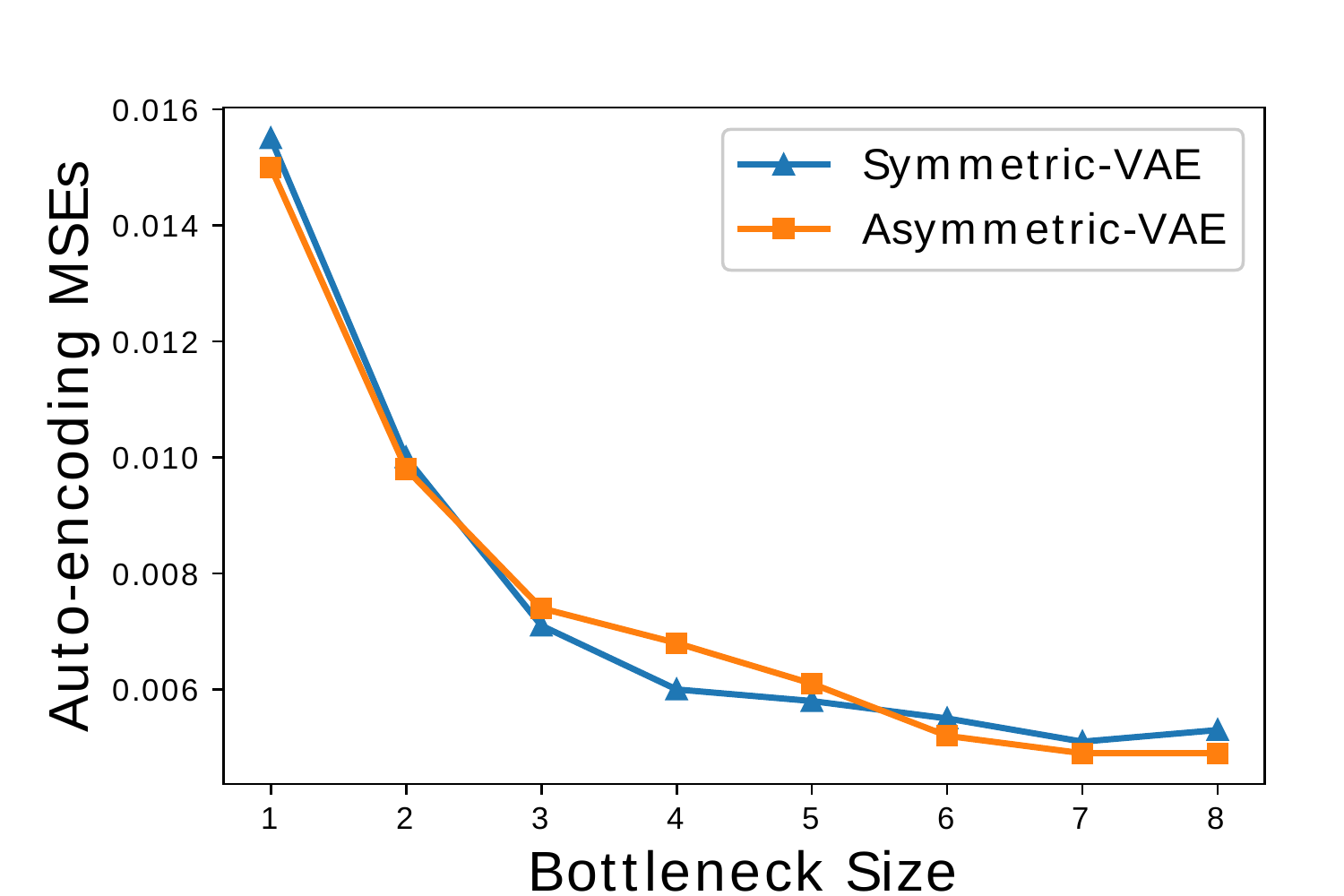}
    \end{subfigure}
    \caption{\label{fig:autoencoding_loss_bottleneck_size}The left figure shows the auto-encoding mean-squared errors (MSEs) of \textbf{Modified LSTM-AE} with different bottleneck sizes for chapter $k=3$, $7$ and $11$. The right figure shows the auto-encoding MSEs of \textbf{Symmetric-VAE} and \textbf{Asymmetric-VAE} with different bottleneck sizes for chapter $k=12$.}
\end{figure*}

\section{Experiments}
In this section, we first list some details of the experimental setups and then present the complete model comparison results.

\subsection{Experimental Setups}
\subsubsection{The Data Set}
We train all of the baseline predictors and unsupervised models on a 12-chapter-long Open edX MOOC, named "Introduction to Computing with Java", which was held by the \textit{Hong Kong University of Science and Technology} from June 2014 to September 2014. There are 44,920 students enrolled but only around one-ninth of them completed the entire course. As described in Sec.\ref{sec_activity_and_assessment}, we prepared 20 features ($F=20$) and 1 label for each student in each chapter. All of the features and labels are rescaled to $[0, 1]$ respectively. The normalization is for each feature/label across all students. For each student, the sequence of features are $[\x_1, \x_2, \dots, \x_N]$, where $N=12$ is the number of chapters. For $1\leq n\leq N$, we have $\x_n\in[0,1]^F$, where $F=20$ is the number of features. The labels also form a sequence $[y_1, y_2, \dots, y_N]$, where each $y_n\in[0, 1]$ (except from $y_{12}$ which is undefined, since there is no assessment in the last chapter of the course). Only students with valid labels at the chapter level are considered. Under such a criterion, our dataset contains 5,739 students. 

\subsubsection{Model Comparison Scheme}
We compare the effectiveness of the learned embeddings of the three unsupervised models by means of the principal component analysis (PCA), the t-distributed stochastic neighbor embedding (t-SNE) \cite{maaten2008visualizing} and the final examination by evaluating the performance improvements on a specific grade prediction.

PCA is a technique that offers the reduction of a large set of correlated variables to a smaller number of uncorrelated hypothetical components \cite{jolliffe1986principal}. It is widely used for feature extraction and selection. The set of output components are ranked in decreasing order of variances, and thus the first several components are most informative. In our experiments, we evaluate the effectiveness of the learned embeddings in the two-dimensional projected space composed of the first two components output by PCA, by visualizing how samples are distributed in the embedding space. If there are clearly formed clusters and our labels are discriminative for distinguishing samples in different clusters, the learned embedding should be discriminative for predicting the label as well. We also compute the percentage of variance retained by the first several components obtained from PCA, to have a sense of how many components are useful and how compact our embeddings could be.

t-SNE as a machine learning algorithm for dimensionality reduction which has also been widely and successfully applied to visualize the learned representations. It is a variation of Stochastic Neighbor Embedding \cite{hinton2003stochastic} that produces visualizations by reducing the tendency to crowd points together in the center of the map. As a nonlinear dimensionality reduction technique, it is particularly well-suited for embedding high-dimensional data into a space of two or three dimensions. In this paper, we use t-SNE as another method to check the effectiveness of the representation learning.

The comparison of prediction performance is carried out as follows. Recall that for each predictor, we trained a sequence of independent models $f = [f_2, \dots, f_N]$. For any $1< k\leq N$, the model $f_k$ of chapter $k$ takes the subsequence $[\x_1, \x_2, \dots, \x_{k-1}]$ as inputs and gives the prediction $\hat y_{k;f} = f_k([\x_1, \x_2, \dots, \x_{k-1}])$. When we compare the prediction performance of two models $f$ and $g$, we actually compare each pair of mean squared errors (MSEs) $\mathrm{MSE}_{f, k}$ and $\mathrm{MSE}_{g, k}$ given a specific chapter $k$. Where $\mathrm{MSE}_{f, k}$ is the average cross-validation MSE of the model $f$ in chapter $k$. In this paper, we always perform five-fold cross-validation and this ensures that the average validation MSEs reflect the actual performance of a model.

\begin{figure*}[t]
    \centerfloat
    \begin{subfigure}[t]{0.31\linewidth}
        \includegraphics[width=\linewidth]{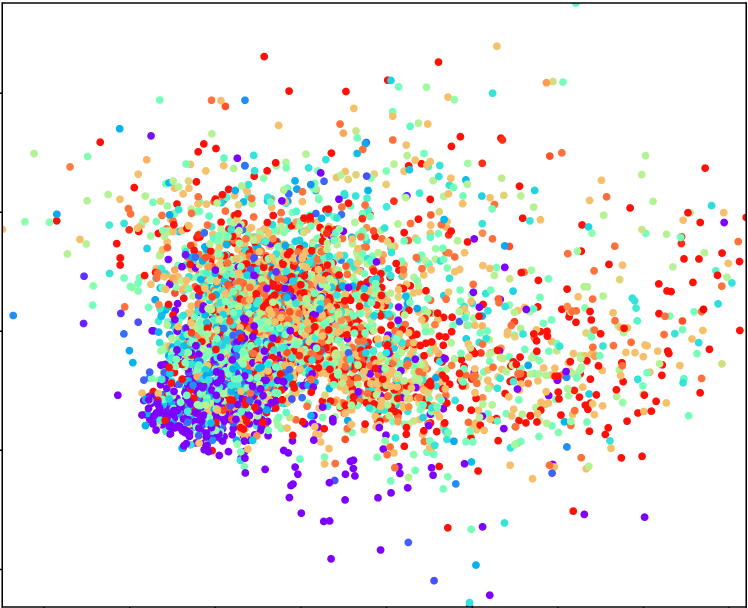}
    \end{subfigure}
    \begin{subfigure}[t]{0.31\linewidth}
        \includegraphics[width=\linewidth]{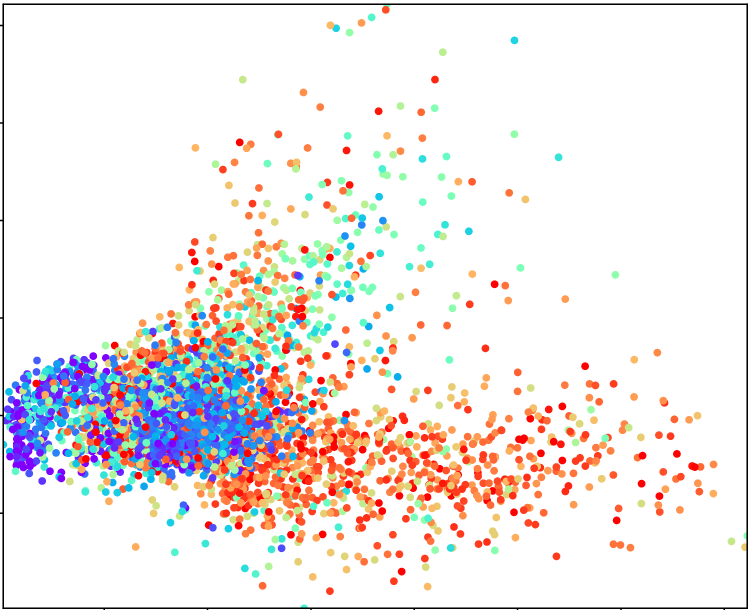}
    \end{subfigure}
    \begin{subfigure}[t]{0.365\linewidth}
        \includegraphics[width=\linewidth]{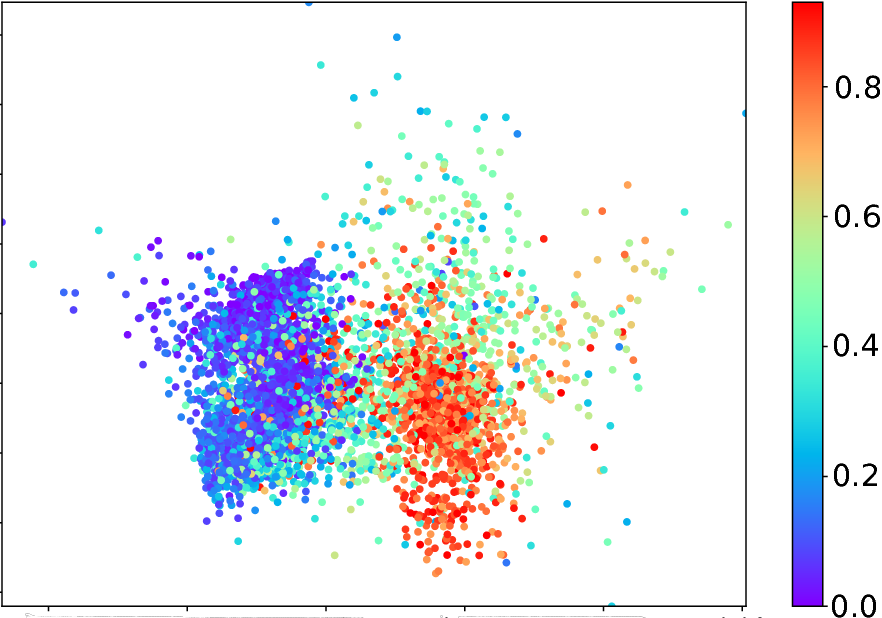}
    \end{subfigure}
    \caption{\label{fig:embedding_pca_chapter} PCA projections of the embeddings learned by \textbf{Modified LSTM-AE}. From left to right, chapter $k$ is $3$, $7$ and $11$ respectively. The color labels the student's average chapter grade up till chapter $k$.}
\end{figure*}

\begin{figure*}[t]
    \centerfloat
    \begin{subfigure}[b]{0.28\linewidth}
        \includegraphics[width=\linewidth]{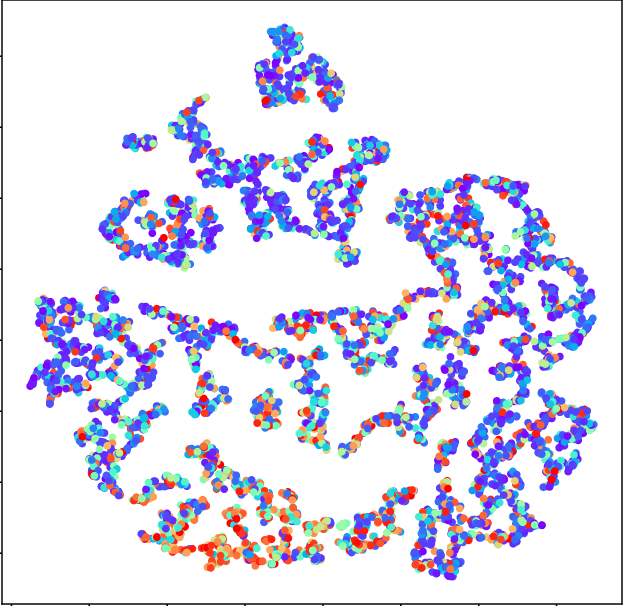}
    \end{subfigure}
    \begin{subfigure}[b]{0.28\linewidth}
        \includegraphics[width=\linewidth]{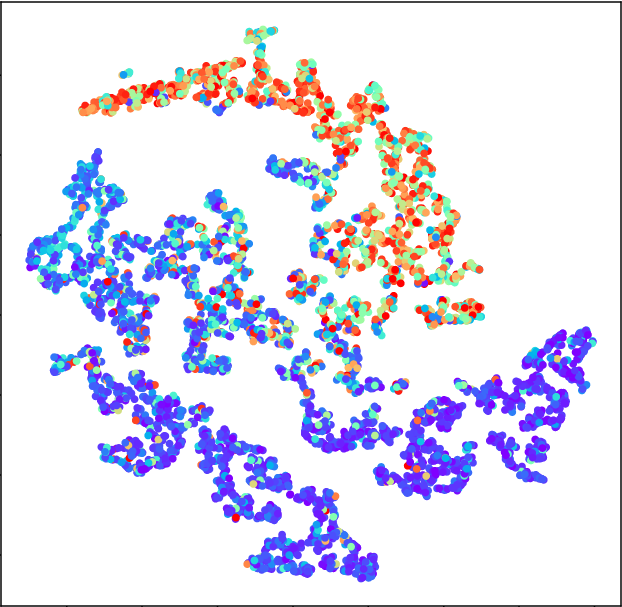}
    \end{subfigure}
    \begin{subfigure}[b]{0.34\linewidth}
        \includegraphics[width=\linewidth]{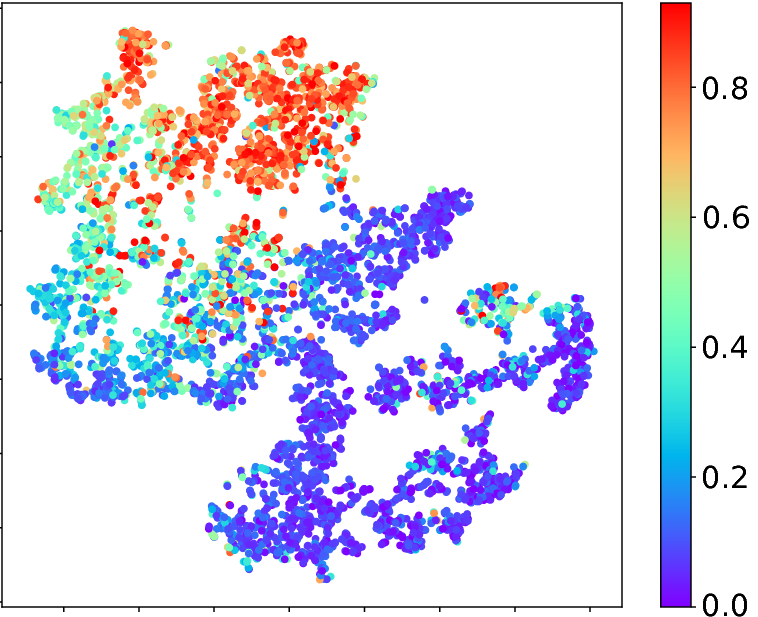}
    \end{subfigure}
    \caption{\label{fig:embedding_tsne_chapter} t-SNE projections of the embeddings learned by \textbf{Modified LSTM-AE}. From left to right, chapter $k$ is $3$, $7$ and $11$ respectively. The color labels the student's average chapter grade up till chapter $k$.}
\end{figure*}

\subsubsection{Experimental Configurations}
Since both our features and labels are scaled to $[0, 1]$, we always use the \textit{sigmoid} function as the output activation function. For the baseline predictors defined in Sec.\ref{sec_prediction_baseline}: \textbf{FC3}, \textbf{CNN2-FC1}, \textbf{LSTM1} and \textbf{CNN1-LSTM1}, we choose a unified learning rate $l=0.001$. While the learning rates for unsupervised models: \textbf{Modified LSTM-AE}, \textbf{Symmetric-VAE}, \textbf{Asymmetric-VAE} are larger $l=0.004$. Following the recommendations in \cite{bengio2012practical}, we increase the number of nodes per layer and the number of epochs until a good fit of the data is achieved. After that, we slightly regularize our network using dropout \cite{srivastava2014dropout} until the model no longer overfits the training data. Here we do not change the number of layers during hyper-parameter tuning, as they are our parameters of interest for model comparison. Since all of our models are not very deep, slight regularization is often adequate and we are able to fit the data well within 200 epochs. We use the \textit{Adam} optimizer \cite{kingma2014adam} for all networks without an LSTM layer, and use \textit{RMSprop} optimizer \cite{tieleman2012lecture} for all LSTM-based models. All networks are implemented using the \textit{Keras} framework with \textit{TensorFlow\texttrademark} back-end.

\begin{figure*}[t]
    \centerfloat
    \begin{subfigure}[b]{0.5\linewidth}
        \includegraphics[width=\linewidth]{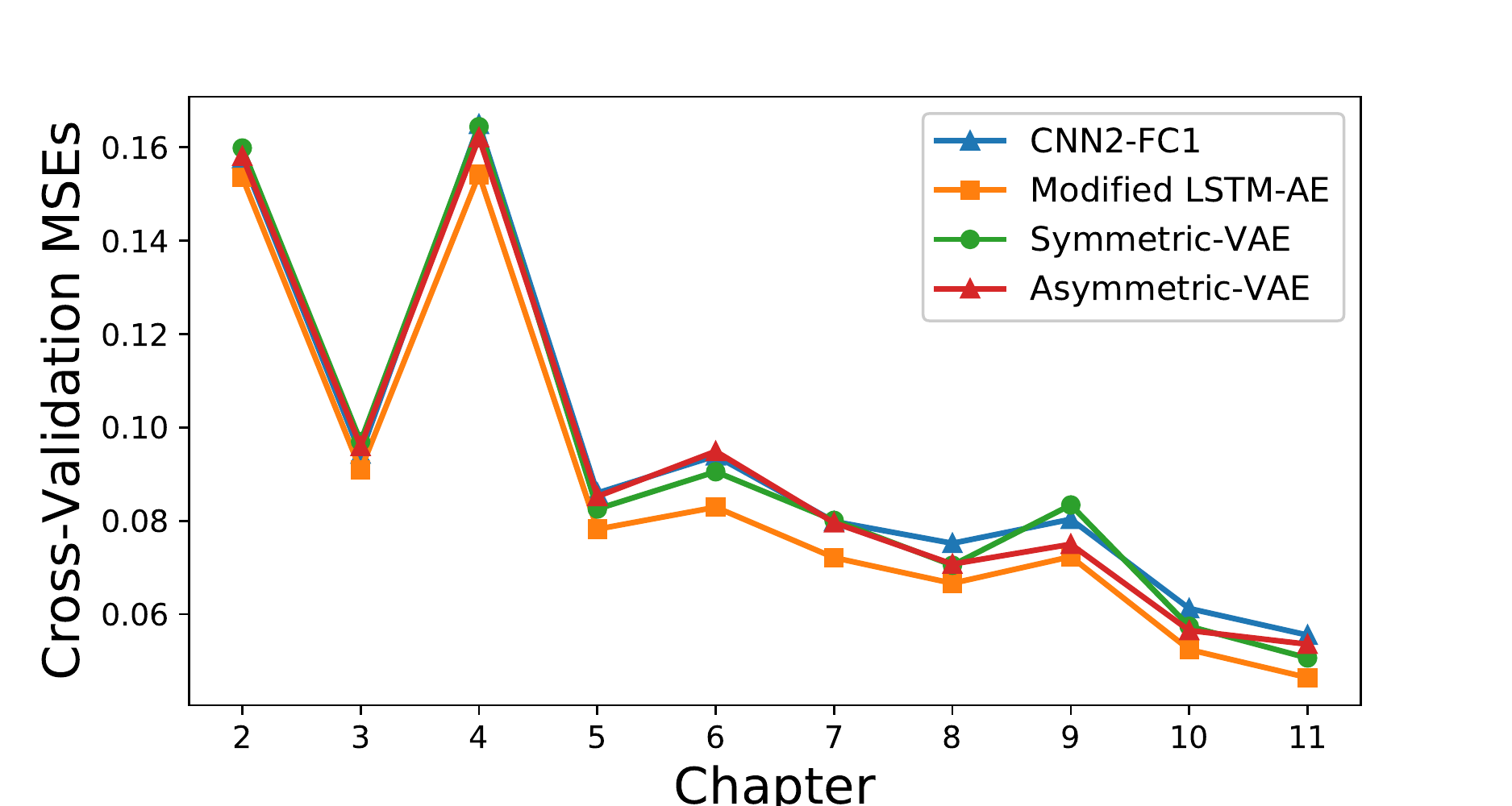}
    \end{subfigure}
    \begin{subfigure}[b]{0.5\linewidth}
        \includegraphics[width=\linewidth]{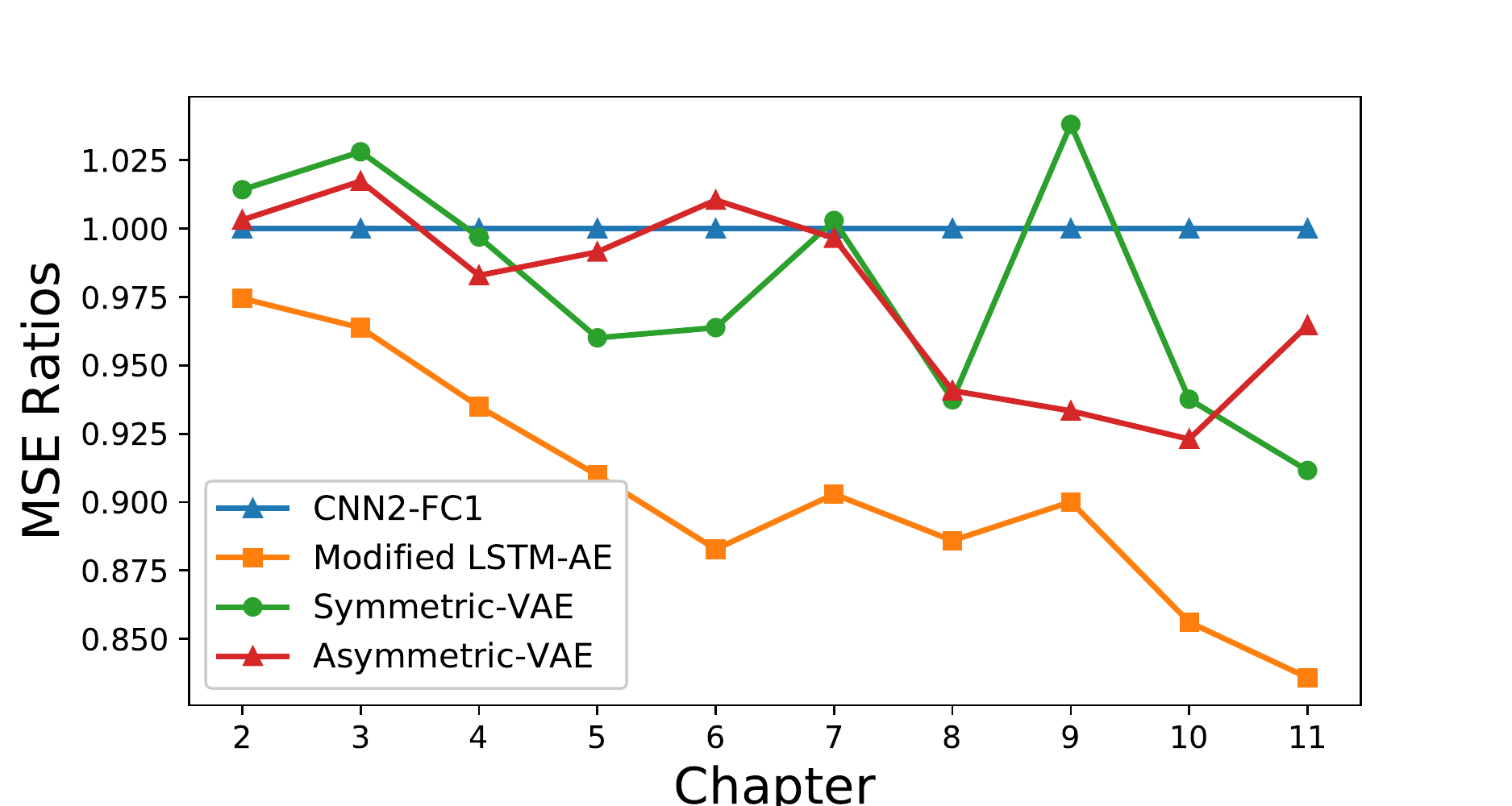}
    \end{subfigure}
    \caption{\label{fig:embedding_predictor_performance} The left-hand side figure shows the difference between prediction performance using either the learned features or the raw features. The right-hand size figure shows the relative prediction performance compared to \textbf{CNN2-FC1} using the raw features.}
\end{figure*}

\begin{figure}[t]
    \centerfloat
    \includegraphics[width=1.1\linewidth]{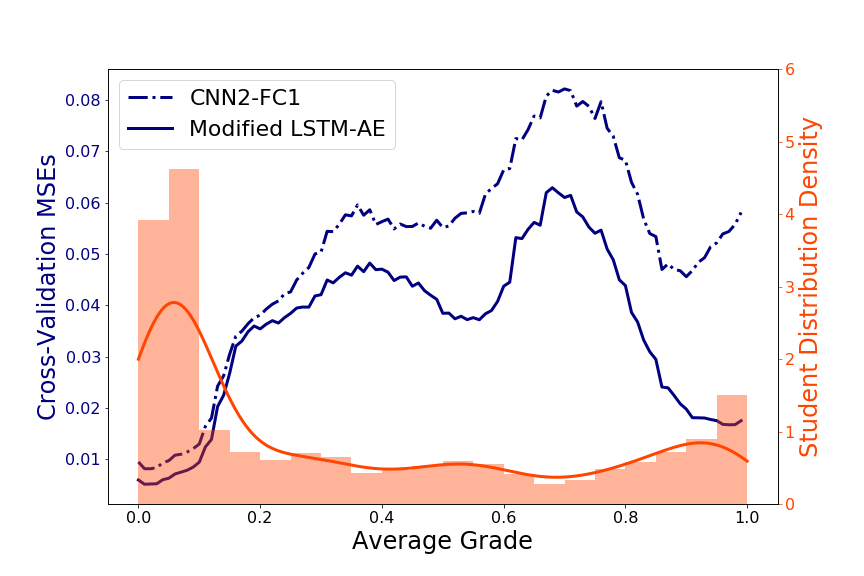}
    \caption{\label{fig:performance_versus_grade} Average prediction performance of \textbf{CNN2-FC1} (blue dashed line) and the predictor using the \textbf{Modified LSTM-AE} embedding (blue solid line) for students with different average grades in all chapters. The three peaks of the distribution density curve (orange line) refer to the low, medium and high performing groups respectively.}
\end{figure}

\subsection{Performance Comparison}
To prove our proposed \textbf{Modified LSTM-AE} model learns a good representation, we compare our approach with the start-of-the-art techniques on finding efficient embeddings (\textbf{Symmetric-VAE} and \textbf{Asymmetric-VAE}) and with the supervised baselines on prediction performance.

\subsubsection{Baseline Predictors}
In the first experiment, we compare the mean squared errors (MSEs) of grade prediction of the baseline predictors. Figure~\ref{fig:predictor_performance} illustrates the relation of average validation MSEs versus the chapter number to predict for all baselines. We can clearly see that neural network models (\textbf{CNN2-FC1}, \textbf{LSTM1} and \textbf{CNN1-LSTM1}) outperform logistic regression (\textbf{LR}) by a large margin. This motivates us using neural networks since traditional regression-based models fail to predict accurately. \textbf{LSTM1} performs weaker than \textbf{CNN2-FC1} when the chapter number $k$ is large, this is because LSTM networks which fit the data best when $k\approx7$ tends to overfit when $k$ is larger. Adding a dimensionality reduction layer (one-dimensional CNN of kernel size 1) (\textbf{CNN1-LSTM1}) solves this problem, where we can see the performance of it and \textbf{CNN2-FC1} are similar for large $k$. We identify that \textbf{CNN2-FC1} is the best predictor architecture having an excellent and stable prediction performance on nearly all chapters. We then choose it as the baseline to compare to predictors working on the embedding representations.

\subsubsection{Bottleneck Sizes}
Before training the unsupervised models, we should carefully choose an important hyper-parameter, the bottleneck size. For a \textbf{Modified LSTM-AE}, the bottleneck size $Z$ is just the dimension of the embedding space $\boldsymbol{z}\in\mathbb{R}^Z$. In Figure~\ref{fig:autoencoding_loss_bottleneck_size}, we plot the cross-validation auto-encoding MSEs of the \textbf{Modified LSTM-AE} model versus the bottleneck size $Z$ on the left-hand side, for three chapters $k=3$, $7$ or $11$. The results show that for the middle chapters $k\approx7$, the MSEs are more sensitive to the bottleneck size. For all situations, setting the bottleneck size to $Z=8$ is enough for retaining its learning capability on our data set. 

Things are completely different for the two VAEs since their embedding features are one-dimensional sequences of size $z\in\mathbb{R}^{Z\times k}$, where $1< k\leq N$, it is expected that a small bottleneck size per unit $Z$ should be enough. When $k=12$ (which is the only situation that they output the entire sequence $X$ as does \textbf{Modified LSTM-AE}), we also plot the curves of auto-encoding MSEs versus bottleneck sizes, as shown in Figure~\ref{fig:autoencoding_loss_bottleneck_size}. We can see that $Z=4$ is generally enough.

\subsubsection{Analysis of the Embedding Representations}
A direct visualization of the embedding representations learned by the \textbf{Modified LSTM-AE} model is shown in Figure~\ref{fig:embedding_pca_chapter} and Figure~\ref{fig:embedding_tsne_chapter}. We can clearly see that at the beginning of a course, the input features are relatively noisy and the learned embedding can hardly separate students with different grades. This situation is improved along with the progression of the course. When $k=11$, we can identify three clusters of students corresponding to the low, medium and high performing groups respectively, with the naked eye. This serves as direct proof that our \textbf{Modified LSTM-AE} model can learn an efficient representation for the chapter grade prediction task.

We also calculate the percentages of variances retained after PCA (when the number of out components is $4$) for all of the three unsupervised models in the case of $k=8$. We keep the sizes of the embedding spaces to be exactly the same so that these ratios are comparable. For the \textbf{Modified LSTM-AE} model, the percentage of variances in the first $4$ components is as high as $97.85\%$, while for the \textbf{Symmetric-VAE} and \textbf{Asymmetric-VAE}, they are $79.84\%$ and $84.72\%$ respectively. A higher retained variance ratio implies a smaller effective size (i.e., the number of important components) of the learned embedding. Thus we conclude that \textbf{Modified LSTM-AE} also learns the most compact representation.

\subsubsection{Predictors on Learned Embedding}
Now to examine how much the prediction performance is improved by the learned embeddings. For each unsupervised model, we train a predictor on a pre-trained encoder and fine-tune the encoder's parameters with a relatively small learning rate (one-tenth of the learning rate on the predictor part of the model). This training methodology is equivalently used for all of the three predictors on embeddings. Figure~\ref{fig:embedding_predictor_performance} illustrates the average validation MSEs for grade prediction in each chapter for the three predictors with different embeddings, and also the results of the supervised baseline, \textbf{CNN2-FC1}. Where we can see the representation learned by \textbf{Modified LSTM-AE} performs the best, as it outperforms the other two VAEs for all chapters by a large margin. The improvements in the early chapters are limited for all of the three unsupervised models because further reducing the dimension of the short feature sequences might remove some useful information that is effective for the prediction problem. We also examine the performance on students with different average grades, as depicted in Figure~\ref{fig:performance_versus_grade} where we can see that the predictor using \textbf{Modified LSTM-AE} embedding features achieves significantly smaller MSEs for the medium and high performing groups, compared with \textbf{CNN2-FC1}. Although these two groups have much smaller numbers of students, the \textbf{Modified LSTM-AE} does not overfit to the dominant low-performing group. It helps reduce the modeling overfit of the supervised methods.
We summarize that \textbf{Modified LSTM-AE} performs consistently and significantly better than the other unsupervised models on improving the real-time grade prediction, where up to 17\% MSEs could be reduced on our data set. 

\section{Conclusion}
In this paper, we report the effective feature learning with unsupervised learning approach for improving the predictive models in MOOCs. In the field of learning analytics, there is a large degree of freedom of designing the handcrafted features for improving the prediction performance in a specific task with an abundance of the clickstream data. Different from those traditional approaches of feature engineering, we attempt to describe students' learning activities in the time-content domain with the raw interaction records in the clickstream. We design the \textbf{Modified LSTM-AE} model which can learn a compact representation that is discriminable for the target labels of performance indicators. In our experiment, the \textbf{Modified LSTM-AE} successfully gives the effective features, which are indeed helpful for improving the prediction performance of the task of predicting students' learning performance, and reducing the modeling overfit to the low-performing majority. An open source release of our pipeline will be published.

For future work, we will examine our approach on more courses when the data is available. Although a specific prediction objective is used to prove that the learned features are effective, our feature learning pipeline is not restricted to a specific supervised learning model nor a specific prediction task. We will also test the representation learning methods on different prediction tasks in the future. To extend our work further, we can attempt other approximations for preparing the feature in the time-content domain. Our features are two-fold in the time domain as a reasonable approximation, some learning patterns in the time domain cannot be identified. A new unsupervised learning approach is needed to learn the effective features that characterize the learning patterns in both the time domain and the content domain.